\pdfoutput=1

\documentclass[conference,compsoc]{IEEEtran}
\usepackage{amsmath} 
\usepackage{caption}
\usepackage{subcaption}
\ifCLASSOPTIONcompsoc
  \usepackage[nocompress]{cite}
\else
  \usepackage{cite}
\fi
\ifCLASSINFOpdf
  \usepackage[pdftex]{graphicx}
  \graphicspath{{png/}}
\else
\fi
\usepackage{fixltx2e}
\begin{document}
\title{Lip Reading Using Convolutional Auto Encoders as Feature Extractor}

\author{
% \IEEEauthorblockN{Dharin Parekh}
% \IEEEauthorblockA{parekh.dharin201@gmail.com}
% \and
% \IEEEauthorblockN{Ankitesh Gupta}
% \IEEEauthorblockA{ankiteshguptas@gmail.com}
% \and
% \IEEEauthorblockN{Shharrnam Chhatpar}
% \IEEEauthorblockA{sharnam19.nc@gmail.com}
% \and
% \IEEEauthorblockN{Anmol Yash}
% \IEEEauthorblockA{anmol100@gmail.com}
% \and
% \IEEEauthorblockN{Prof. Manasi Kulkarni}
% \IEEEauthorblockA{mukulkarni@ce.vjti.ac.in}
\makebox[.33\linewidth]{Dharin Parekh\textsuperscript{1}}\\parekh.dharin201@gmail.com\\
  \and \makebox[.33\linewidth]{Ankitesh Gupta\textsuperscript{1}}\\ankiteshguptas@gmail.com\\
  \and \makebox[.33\linewidth]{Shharrnam Chhatpar\textsuperscript{1}}\\sharnam19.nc@gmail.com\\
  \and \makebox[.5\linewidth]{Anmol Yash\textsuperscript{1}}\\anmol100@gmail.com\\
  \and \makebox[.5\linewidth]{Prof. Manasi Kulkarni\textsuperscript{1}}\\mukulkarni@ce.vjti.ac.in\\
  \and \makebox[1.0\linewidth]{\textsuperscript{1}CE \& IT Department, Veermata Jijabai Technological Institute, Mumbai}
}


% compsoc conferences compared to traditional conferences), use this
%Telephone: (800) 555--1212, Fax: (888) 555--1212}


\maketitle

\begin{abstract}
Visual recognition of speech using the lip movement is called Lip-reading. Recent developments in this nascent field uses different neural networks as feature extractors which serve as input to a model which can map the temporal relationship and classify. Though end to end sentence level Lip-reading is the current trend, we proposed a new model which employs word level classification and breaks the set benchmarks for standard datasets. In our model we use convolutional autoencoders as feature extractors which are then fed to a Long short-term memory model. We tested our proposed model on BBC's LRW dataset \cite{LRW}, MIRACL-VC1\cite{MIRACL} and GRID\cite{GRID} dataset. Achieving a classification accuracy of 98\% on MIRACL-VC1 as compared to 93.4\% of the set benchmark by \cite{Rekik}. On BBC's LRW the proposed model performed better than the baseline model of convolutional neural networks and Long short-term memory model as seen in \cite{Garg}. Showing the features learned by the models we clearly indicate how the proposed model works better than the baseline model. The same model can also be extended for end to end sentence level classification.
\end{abstract}


% \fi
\IEEEpeerreviewmaketitle

\section{Introduction}

Lip-reading is a technique of understanding speech by visually interpreting the movements of the lips, face and tongue. In noisy environments, where speech recognition is difficult, visual speech recognition offers an effective way to understand speech. Lipreading is a challenging problem due to the different accents, speed of speaking, facial features, skin color etc. However, there are a host of applications, due to which this problem assumes significance. It is immensely helpful for the hearing impaired, assists in understanding spoken language in a noisy environment etc.

\IEEEPARstart
The task of lip reading primarily consists of two processing block. The first block is responsible to extract relevant features from the input video frame, while the other models the relationship between the features of these video frames. The task of the first block becomes tedious as many systems use very complex or manual methods to extract the features. Although sometimes the whole system performs good, but it is quiet impractical to perform manual feature extraction.

\IEEEPARstart
Several works in this field have been proposed in the recent years, which primarily uses neural networks to classify the utterances. These systems also uses neural networks based techniques to preprocess the dataset and to extract relevant features, but many a times these techniques fail to explain what features the method has learned, in-turn affecting the overall accuracy.

\IEEEPARstart
In this work we propose the use of Convolutional Autoencoders (CAE) as defined in \cite{CAE} to extract lip features from the video frames, these features are then given as an input to the Long Short
Term  Memory (LSTM;\cite{LSTM}) which gives the final trained model. We have taken conventional Convolutional Neural Networks (CNN;\cite{CNN}) to extract features as our baseline model, the features learned by this model is compared with our proposed model in terms of the convolved input images.

\IEEEPARstart
The remaining of this paper is organized as follows. First we describe the multiple datasets used to test our network and the pre-processing methods performed on those. Then we define the architectures used to compare the results, firstly we state our baseline model and then we explain our proposed model. Finally in the experimentation section we first compare the features learned with the respective models and later we state the results of our experiments.

\section{Related Works}
 With the advent of access to powerful GPUs, solving problems using deep learning has become quite popular as they are producing state of the art results. In this section, we enlist various existing approaches to automated lip reading using deep learning.

\IEEEPARstart
Chung \& Zisserman (2016a) in \cite{LRW} used variations of convolutional neural networks for word-level classification. They added extra layers(convolution, pooling) above the VGG-M architecture,  the convolutional layer used 3D convolution operator to convolve the input image. The two variations namely Early Fusion(EF) and Multi Tower(MT) captured the spatial and spatiotemporal aspect of the input data respectively. The network was trained on the BBC dataset \cite{LRW}  and achieved an accuracy of 61.1\% on a vocab size of 500 words.

\IEEEPARstart
Garg et al in \cite{MIRACL} used pre-trained VGGNet which were trained on faces, they retained most of the VGGNet and only replaced the final fully connected layer to meet their classification needs. They used the MIRACL-VC1 dataset \cite{MIRACL} for their word-level classification task.
Primarily, they changed the input data that is being fed to the model, they concatenated the whole video to make it one image, also normalizing the speed of each speaker by considering total frames to be 25. The model achieved an accuracy of 56\% (speaker independent) on a vocab size of 10 words. Rekik et al in \cite{Rekik} used hidden Markov models (HMM) to predict utterances, they achieved an accuracy of 93.4\% in a speaker dependent setting and 62.1\% in speaker independent.

\IEEEPARstart
Wand et al in \cite{Wand} proposed a joint architecture where fully connected feed-forward networks were used to extract features of the input frames which was in turn fed to LSTM to model sequential dependency. They used GRID corpus \cite{GRID} with a vocab size of 51 words. Their model achieved an accuracy of 79.6\% in a speaker dependent setting.

\section{Datasets \& Preprocessing}
In this section, we discuss the chosen datasets and video preprocessing methods. We make use of multiple datasets in the paper which are BBC's LRW\cite{LRW} dataset, GRID\cite{GRID} a sentence level dataset and MIRACL-VC1\cite{MIRACL} a dataset consisting of words and phrases.

\subsection{Characteristics}

\IEEEPARstart
LRW\cite{LRW} dataset originally consists of 500 words each word having 1000 occurrences, for the purpose of this paper we make use of two subsets of the LRW dataset due to it's large size, one consisting of only 9 words with all 1000 occurrence of each word. We refer to this subset as BBC-9. Another subset from the LRW consisted of 27 words with all 1000 occurrence of each word. We refer to this as BBC-27. The main difference between BBC-9 and BBC-27 is that the 9 words in BBC-9 are simpler words while BBC-27 consists of confusing words. The simpler words are chosen in such a way that each word starts with a different letter of the english alphabet, and the confusing words were selected such that two or more words might exist that start with the same letter of the english alphabet. The train, val and test set of BBC-9 and BBC-27 consists of 900, 50 and 50 occurrences of each word. 

\IEEEPARstart
GRID\cite{GRID} dataset is a sentence level dataset comprising of utterances from 34 speakers, due to it's large size our experiment was based on only utterances of 5 speakers with each video segmented on words so as to form word level dataset. The split of train, val and test set was set to 90\%, 5\% and 5\% of each speaker respectively. Henceforth, we refer to this dataset as GRID-5. 

\IEEEPARstart
In MIRACL-VC1\cite{MIRACL} words dataset which has a vocabulary of 10 words, there are 15 speakers each speaking a word 10 times. We perform two kinds of experiments on MIRACL-VC1, speaker dependent testing and speaker independent testing\cite{Rekik, Abiel, Garg}. Dataset for speaker dependent testing was formed using 8 random occurrences of all words for each speaker in the training set. The remaining 2 occurrences are distributed across val and test set. We call this dataset as Miracl-Speaker-Dependent (MSD) dataset. The speaker independent testing was performed 15 times, each test included a different speaker in the test set. Of the remaining 14 speakers, 13 were added to train and 1 to the val set. We collectively refer to these as Miracl-Speaker-Independent (MSI) dataset. For MSI dataset, we report the average accuracy over the 15 tests performed.

% \begin{table}[!t]
% % increase table row spacing, adjust to taste
% \renewcommand{\arraystretch}{1.3}
% %if using array.sty, it might be a good idea to tweak the value %of
% %\extrarowheight as needed to properly center the text within the %cells
% \caption{Datasets Table}
% \label{dataset_table}
% \centering
% % Some packages, such as MDW tools, offer better commands for making tables
% % than the plain LaTeX2e tabular which is used here.
% \begin{tabular}{|c|c|c|c|c|c|}
% \hline
% \textbf{ } & \textbf{A} & \textbf{B} & \textbf{C} & \textbf{D} & \textbf{E}\\
% \hline
% \textbf{Source} & BBC & BBC & GRID & MIRACL-VC1 & MIRACL-VC1\\
% \hline
% \textbf{Vocabulary} & 9 & 27 & 51 & 10 & 10\\
% \hline
% \textbf{Train} & 8100 & 24300 & 5 speakers & 10 & 10\\
% \hline
% \end{tabular}
% \end{table}

\subsection{Preprocessing}
\label{normal}
 We used pretrained haar cascade of OpenCV\cite{Bradski} to extract the mouth region from each frame. Each frame after extracting the mouth region is as shown in Figure 1b. Gray scaled images are used instead of coloured frames to reduce the number of features as color frames have three channels which increases the number of features three times. To make the number of frames constant in every video, black frames are appended. The constant number of frames is derived from the dataset's characteristics, for BBC's LRW dataset\cite{LRW} the number of frames are $29$ whereas for MIRACL-VC1\cite{MIRACL} and GRID\cite{GRID} it is $25$.
\begin{figure}[htbp]
    \centering
    \begin{subfigure}[b]{0.25\textwidth}
      \centering
      \includegraphics[width=1.5in]{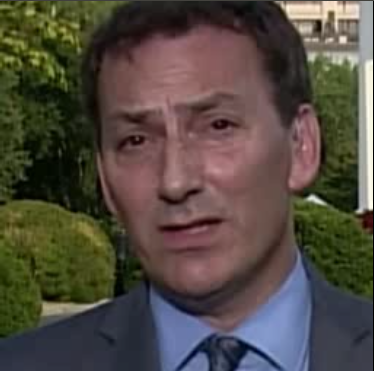}
        \caption{Original frame in BBC's LRW dataset\cite{LRW}}
    \end{subfigure}
    \hfill
    \begin{subfigure}[b]{0.25\textwidth}
      \centering
      \includegraphics[width = 1.5in]{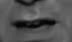}
        \caption{Preprocessing for one frame of BBC's LRW\cite{LRW} dataset}
    \end{subfigure}
    \caption{Preprocessing for one frame of BBC’s LRW \cite{LRW} dataset}
    \label{fig:preprpcessing}
\end{figure}

% \subsection{Data Augmentation}
% We make use of data augmentation techniques in case of MIRACL\cite{MIRACL}, because of less training examples. We increase the dataset by flipping the image horizontally, and shifting both the flipped image and original image by 1 pixel in either horizontal or vertical direction or both.

\section{Architectures}
In this section, we define our architecture and compare it with the baseline model in terms of the features learned. The architectures were first tried on the  BBC's LRW dataset after preprocessing the data as defined in section \ref{normal}, once we achieved satisfying results, the model was then tested with other datasets too.

\subsection{Baseline Model}
\begin{figure}[h!]
    \includegraphics[width=\columnwidth ,height=170pt]{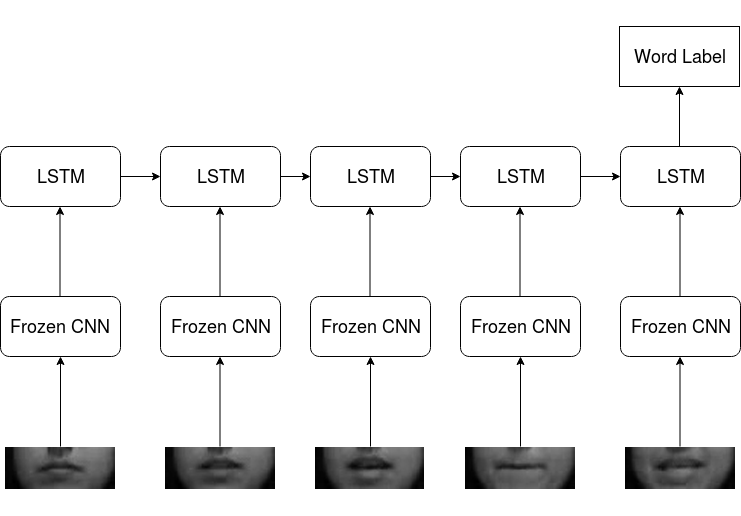}
    \caption{Baseline model}
    \label{fig:baseline}
\end{figure}

Our baseline model was inspired by the one defined in \cite{Garg}. In that paper, the authors used CNNs\cite{CNN} as feature extractor which was then fed to LSTMs\cite{LSTM} to model the objective function. They used pre-trained frozen VGGNet\cite{VGGNET} which was trained on human faces. We used the same concept, but we trained a different CNN architecture, which was frozen and the extracted features were given to LSTMs as their input whose output was in-turn converted into probabilities to classify as words. Our CNN was pre-trained on the images of the speaker's lips. Figure \ref{fig:baseline} describes the model pictorially.

Consider X to be the dataset containing images of lips and nonlips which was created by taking random patched from video frames. The CNN was initially trained to classify images in X(using softmax function to calculate probabilities) as shown in the below given equation. Once the model was trained the softmax layer was removed.

\label{cnnEqn}
\begin{align*}
Output &= CNN(X) \\ 
Probabilities  &= softmax(Output) 
\end{align*}
\newline
Assume x to be an input video(sequence of 29 frames), the frame at time t was fed to the trained CNN's last layer and the output was gathered and was fed to the LSTM as shown:

\begin{align*}
    features &= gather(\forall{t_{0\rightarrow{29}}} CNN(x_{t} )) \\
    label &= LSTM(features)
\end{align*}
\newline
This model's results were taken as the baseline for our architecture and are compared in Section \ref{results_baseline}.

\subsection{CAE + LSTM}
\begin{figure}[h!]
    \includegraphics[width=\columnwidth]{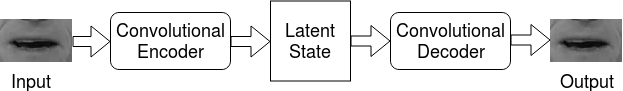}
    \caption{Convolutional Autoencoder}
    \label{fig:encoder_decoder}
\end{figure}

In this model, we make use of a CAE to extract image features. For this model we perform training in two phases. In the first phase of training, we train the CAE. Consider X to be the dataset containing extracted regions of lip using Haar Cascade. The CAE was trained to reproduce the input as shown in the equation below. Figure \ref{fig:encoder_decoder} describes the autoencoder model pictorially.
\label{autoencEqn}
\begin{align*} 
Encoder Output = ConvolutionalEncoder(X) \\
X = ConvolutionalDecoder(Encoder Output)
\end{align*}

\begin{figure}[h!]
    \includegraphics[width=\columnwidth]{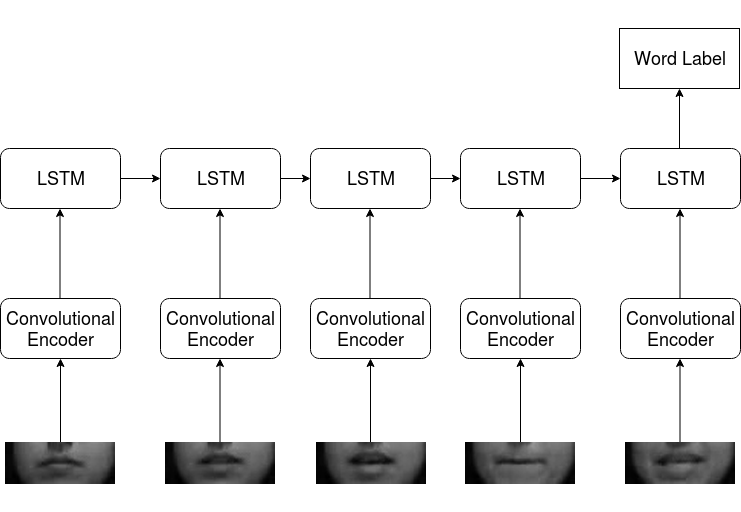}
    \caption{Proposed model CAE (Convolutional AutoEncoder) + LSTM}
    \label{fig:CAE_model}
\end{figure}

Once the CAE model is trained, we discard the convolutional decoder, and only use the convolutional encoder as a feature extractor for images. In the second phase of training, we pass each frame of the video through this trained convolutional encoder to get their features. Once we have obtained the features for each frame we begin training the LSTM, as shown in Figure \ref{fig:CAE_model}. In this step, we forward the time series features of the videos to the LSTM Classifier for training. The LSTM classifies the word based on the features it received as input. The equation for training LSTM is shown in the below equation.

\begin{align*} 
features &= gather(\forall{t_{0\rightarrow{29}}}ConvolutionalEncoder(x_{t}))\\
label &= LSTM(features)
\end{align*}

\section{Experimentation}
In this section we elaborate the experimental setup and also compare the features learned by our proposed model and the baseline model. We also compare the results achieved by our model with other state of the art models.
\subsection{Experimental Setup}
Once the preprocessing was done, different datasets were cropped to different dimensions to satistfy the model's need, Table \ref{tab:dimension_table} shows the mapping of a dataset to that of its dimension.
\begin{table}[h!]
    \renewcommand{\arraystretch}{1.5}
    \caption{Dimensions of frames in different datasets}
    \centering
    \begin{tabular}{|c|c|c|c|}
    \hline
    \textbf{ } & \textbf{BB's LRW} & \textbf{MIRACL-VC1} & \textbf{GRID}  \\
    \hline
    \textbf{Dimension (WxH)} &  72x42 & 72x28 & 72x28 \\
    \hline
    \end{tabular} \\
    \label{tab:dimension_table}
\end{table}

For the BBC dataset, baseline model was trained using a 5 layered CNN with increasing number of kernels. To reduce the dimension of the image after convolution operation max pooling was used. After the series of convolution-activation-pooling operations a hidden layer with 100 nodes was added after which softmax operation was performed. Here 100 is basically the feature dimension of the input frame, this was further fed to LSTM. LSTM had 512 nodes as its hidden dimension, weights of every parameter were initialized using Xavier initialization as described in \cite{Xavier}.

\begin{figure}
    \centering
    \begin{subfigure}[b]{0.25\textwidth}
       \centering
       \includegraphics[width=1.5in]{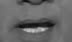}
        \caption{Original input image}
    \end{subfigure}
    \hfill
    \begin{subfigure}[b]{0.25\textwidth}
       \centering
       \includegraphics[width=1.5in]{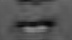}
        \caption{Decoded image}
    \end{subfigure}
    \caption{Generated lip image from the hidden representation of the input frame}
    \label{fig:caeOutput}
\end{figure}

To build the CAE the architecture of CNN was similar to that of baseline's CNN. To train CAE mean squared metric was used, all the models were programmed using tensorflow(\cite{tensorflow}). Figure \ref{fig:caeOutput} shows the original image and the decoded image of the original image when it was fed to the CAE. The blurriness of the decoded image can be attributed to non-uniform distribution of the dataset.

\subsection{Learned feature comparison} \label{ssec:learnedcomparision}

\begin{figure}[h!]
  \begin{subfigure}[b]{0.4\textwidth}
    \includegraphics[width=\textwidth]{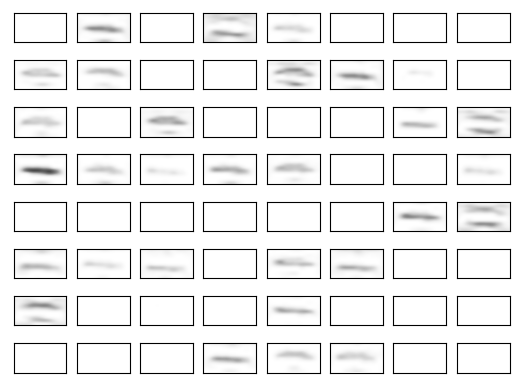}
    \caption{Baseline}
    \label{fig:baselineFeatures}
  \end{subfigure}
  \begin{subfigure}[b]{0.4\textwidth}
    \includegraphics[width=\textwidth]{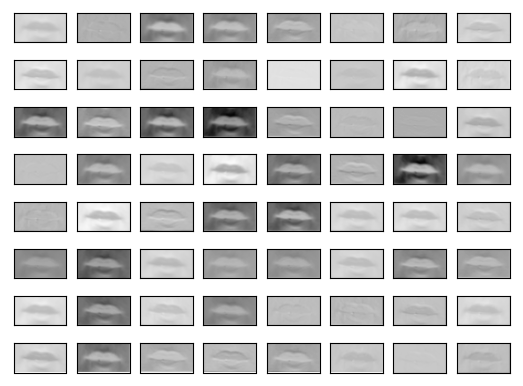}
    \caption{CAE}
    \label{fig:CAEfeatures}
  \end{subfigure}
  \caption{Features learned by the baseline model and the CAE model}
\end{figure}

To visualize the features learned by the CNN model the values of 1st layer's kernels were convolved with the input image frame to represent the features learned. There were 64 kernels in the first convolutional layer, which are expected to learn some representation of the input image. The baseline model learns various aspects of the input image as show in Figure \ref{fig:baselineFeatures}, it is quiet visible that insignificant number of kernels learns the features, as many of the convoluted images are empty. On the contrary, CAE model learns a significant number of features as depicted in  Figure \ref{fig:CAEfeatures}, as very few convoluted input frames are empty.

\subsection{Results}
In this section we try to compare the results of our proposed model with several other state of the art models on different datasets. At places where we couldn't find any existing work on a particular type of dataset we have compared the results with our baseline model. The metric of comparison we will be using is classification accuracy. Only the approaches with the best results are mentioned.

\subsubsection{BBC's LRW} \label{results_baseline}
When CAE + LSTM model was fed the BBC-9 dataset, the results that we got on the test set being 85.61\% were much better than the results on our baseline model result of 79.45\% . The reason for this increase in classification accuracy could be attributed to the kernel features that were learned by the models as shown in Section \ref{ssec:learnedcomparision}

\begin{table}[!th]
\renewcommand{\arraystretch}{1.5}
\caption{Results table of Baseline Model and CAE + LSTM model on BBC-9 dataset}
\label{tab:baselinevscaeresults}
\centering
\begin{tabular}{|c|c|c|c|}
\hline
\textbf{ } & \textbf{train} & \textbf{val} & \textbf{test}\\
\hline
\textbf{Baseline Model} & 87.47\% & 82.13\% & 79.45\%\\
\hline
\textbf{CAE + LSTM} & 97.91\% & 89.79\% & 85.61\%\\
\hline
\end{tabular}
\end{table}

\begin{figure}[h!]
    \includegraphics[width=\columnwidth]{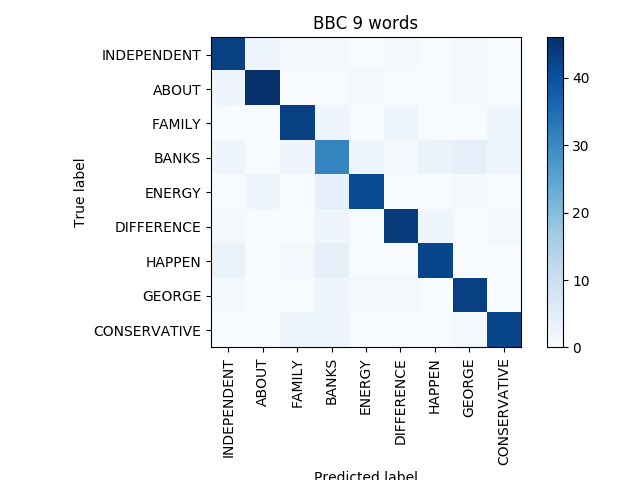}
    \caption{Confusion matrix of BBC-9}
    \label{fig:confusion9}
\end{figure}

\begin{figure}[h!]
    \includegraphics[width=\columnwidth]{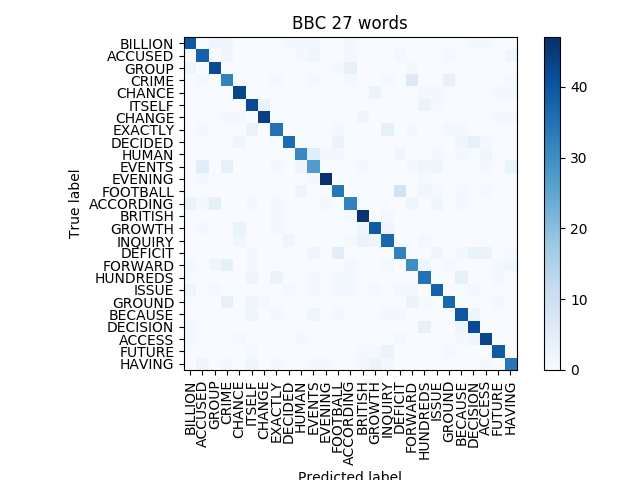}
    \caption{Confusion matrix of BBC-27}
    \label{fig:confusion27}
\end{figure}

As we can refer from Table \ref{tab:baselinevscaeresults} the results on the CAE + LSTM model were significantly better than the baseline model, we tried the same model on BBC-27, where we fed 27 confusing words having similar set of phonemes. An example of confusing words is chance and change. Even though the result shows that there is a slight decrease in classification accuracy as compared to BBC-9, the model gives acceptable results even when given similar and confusing words. The results of CAE + LSTM on BBC-27 can be observed from Table \ref{tab:bbc27caeresults}. The confusion matrix of both BBC-9 and BBC-27 as shown in Figure \ref{fig:confusion9} and in Figure \ref{fig:confusion27} respectively shows the highest intensity along the diagonal which indicates the on par performance of our proposed model.

\begin{table}[!th]
\renewcommand{\arraystretch}{1.5}
\caption{Results table of encoder decoder model on BBC-27 dataset}
\label{tab:bbc27caeresults}
\centering
\begin{tabular}{|c|c|c|c|}
\hline
\textbf{ } & \textbf{train} & \textbf{val} & \textbf{test}\\
\hline
\textbf{CAE + LSTM} & 86.75\% & 73.74\% & 77.421\%\\
\hline
\end{tabular}
\end{table}

\subsubsection{MIRACL-VC1}
On MIRACL-VC1 as described earlier there are two types of dataset which we refer as Miracl-Speaker-Dependent (MSD) and Miracl-Speaker-Independent (MSI). On MSD dataset CAE + LSTM got a test set classification accuracy of 98\% which was a improvement than the 93.4\% obtained in \cite{Rekik}. The results can be seen in Table \ref{tab:miracldependentcompare}.

\begin{table}[!th]
\renewcommand{\arraystretch}{1.5}
\caption{Results table of CAE + LSTM and HOG\textsubscript{c} + HOG\textsubscript{d} + MBH\cite{Rekik} on MSD dataset}
\label{tab:miracldependentcompare}
\centering
\begin{tabular}{|c|c|c|c|}
\hline
\textbf{ } & \textbf{train} & \textbf{val} & \textbf{test}\\
\hline
\textbf{HOG\textsubscript{c} + HOG\textsubscript{d} + MBH \cite{Rekik}} & - & - & 93.4\%\\
\hline
\textbf{CAE + LSTM} & 98.85\% & 97\% & 98\%\\
\hline
\end{tabular}
\end{table}

Considering the MSI dataset, which was a challenging task, as even the state of the art got an accuracy of 62.1\% \cite{Rekik}, we managed to cross that benchmark with a 63.22\% test set classification accuracy as seen in Table \ref{tab:miraclindependentcompare}. The model could have performed better had we followed the 14-1 train-test split as used by \cite{Rekik}, as compared to 13-1-1 train-val-test split which was used by us to keep consistency across all our datasets.

\begin{table}[!th]
\renewcommand{\arraystretch}{1.5}
\caption{Results table of CAE + LSTM and HOG\textsubscript{c} + HOG\textsubscript{d} + MBH\cite{Rekik} on MSI dataset}
\label{tab:miraclindependentcompare}
\centering
\begin{tabular}{|c|c|c|c|}
\hline
\textbf{ } & \textbf{train} & \textbf{val} & \textbf{test}\\
\hline
\textbf{HOG\textsubscript{c} + HOG\textsubscript{d} + MBH \cite{Rekik}} & - & - & 62.1\%\\
\hline
\textbf{CAE + LSTM} & 92.29\% & 59.01\% & 63.22\%\\
\hline
\end{tabular}
\end{table}

\subsubsection{GRID}
To evaluate the performance of our proposed model on another standard dataset, we evaluated the results on GRID-5 dataset, where in the data was close to around 30,000 videos comparable to BBC-27 dataset's size, the accuracy metric shows a significant improvement as compared to BBC-27 test set classification accuracy. The results on GRID-5 dataset can be seen in Table \ref{tab:grid5results}.

\begin{table}[!th]
\renewcommand{\arraystretch}{1.5}
\caption{Results table of CAE + LSTM on GRID-5 dataset}
\label{tab:grid5results}
\centering
\begin{tabular}{|c|c|c|c|}
\hline
\textbf{ } & \textbf{train} & \textbf{val} & \textbf{test}\\
\hline
\textbf{CAE + LSTM} & 86.70\% & 86.37\% & 84.80\%\\
\hline
\end{tabular}
\end{table}

\section{Conclusion}
In this paper we have proposed a model for automated lip reading of words, using only visual input of speakers facial expressions. \\
The proposed model known as CAE + LSTM uses CAE(Convolutional Auto Encoders) as feature extractors and then the temporal data of features is fed to LSTM to get the final classification of words. There are two separate stages for training this model. First is to train the CAE separately to learn the optimal hidden representation of the input frame, second is to train the whole model by using this frozen CAE as a feature extractor.\\
We evaluated our model using the classification accuracy on multiple datasets. For BBC's LRW \cite{LRW} we compared our system with the baseline models classification accuracy. After seeing a significant improvement we evaluated the model on two other standard dataset's. On MIRACL-VC1 speaker dependent testing we crossed the benchmark of 93.4\% by \cite{Rekik} with an accuracy of 98\%. Also on speaker independent testing our accuracy of 63.22\% surpasses the benchmark of 62.1\% by \cite{Rekik}. To evaluate our performance on another standard dataset, we also got 84.8\% test accuracy on GRID dataset with a dataset of 30,000 words video.
Thus the proposed model can be a better approach than CNN's + LSTM and HMM's for automated lip reading, and with the right data and resources the model can even we used for end-to-end lip reading of phrases.



%\bibliographystyle{IEEEtran}

\end{document}